\documentclass{article}

\usepackage[final]{corl_2026}

\usepackage{amsmath}
\usepackage{amssymb}
\usepackage{graphicx}
\usepackage{wrapfig}

\title{Koopman DCM: Unstable Eigenfunctions as Data-driven Representations for Legged Balancing}

\author{
  Stéphane Caron \\
  Institute for Intelligent Systems and Robotics \\
  CNRS and Sorbonne University,
  Paris, France\\
  \texttt{stephane.caron@isir.upmc.fr} \\
}

\begin{document}
\maketitle

%===============================================================================

\begin{abstract}
    In legged locomotion, divergent components of motion (DCMs) have emerged as characteristic states for balance control. They isolate the unstable mode of the dynamics but, in existing formulations, apply only to reduced models such as the linear inverted pendulum. In this study, we show how DCMs can be more generally formulated as Koopman eigenfunctions. Whereas Koopman analysis typically targets eigenvalues near zero, which capture conserved or slowly varying quantities, our investigation leads us to deliberately search for unstable eigenpairs with large eigenvalues. The resulting Koopman DCMs are data-driven observables trained using only real-robot data. On a real biped, DCMs learned from one hour of robot data improve tracking of reference walking patterns. We further show how learned DCMs provide state-based viability constraints when combined with model predictive control.
\end{abstract}


%===============================================================================

\section{Introduction}

Following the walking and stair-climbing demonstrations of the Honda humanoid robot in 1998~\citep{hirai1998development}, legged locomotion control has long focused on \emph{reduced models}, that is, low-dimensional templates of the robot's dynamics that admit tractable control laws. The most widely used is the linear inverted pendulum~\citep{kajita2001lipm}, whose natural control coordinate is the divergent component of motion (DCM)~\citep{takenaka2009real}. The DCM isolates the unstable mode and reduces balance to the regulation of a first-order linear time-invariant system. Yet its derivation is specific to the reduced model. In this work, we lift this restriction by redefining the DCM as a Koopman eigenfunction of the robot's dynamics, and learning it directly from data on the target system.

This study proceeds in four steps. First, we cast the DCM as the solution of an eigenfunction search problem for the Koopman operator, where we seek an unstable eigenvalue. Second, we propose a training pipeline that learns Koopman DCMs from closed-loop measurement-action data. Third, we deploy the learned DCM on a real robot with proportional feedback, and observe that it improves tracking compared to existing DCMs. Finally, we show how learned DCMs can be used to formulate hard viability constraints in model predictive control, which we evaluate in simulation.\footnote{Source code and data are available at \url{https://github.com/stephane-caron/koopman_dcm}.}

%===============================================================================

\section{From DCMs to Koopman Eigenfunctions}

\subsection{The DCM of the Linear Inverted Pendulum}
\label{sec:lip}

The most studied reduced model, the linear inverted pendulum (LIP)~\citep{kajita2001lipm}, rests on two key assumptions: conserved angular momentum around the center of mass (CoM), and a constant CoM height above ground. Under these assumptions, the articulated equations of motion of a legged robot reduce to:
\begin{equation}
  \ddot{c} = \omega^2 (c - z), \qquad \omega := \sqrt{g/h},
  \label{eq:lip}
\end{equation}
where $c \in \mathbb{R}^2$ is the horizontal CoM position, $z \in \mathbb{R}^2$ is the zero-tilting moment point (ZMP), $g \approx 9.81\,\mathrm{m}/\mathrm{s}^2$ is standard gravity, and $h$ is the altitude of the CoM above ground. The ZMP is defined as the intersection of the central axis of the contact wrench with the horizontal ground plane~\citep{sardain2004forces}. Intuitively, it is the point on the ground where the resultant of contact forces applies. The appeal of the LIP model is that it yields a linear time-invariant system with control input $z$. Since contact forces are unilateral and limited by friction, the ZMP is constrained to a convex polygon, the ZMP support area $\{ z | C z \leq d\}$~\citep{caron2016tro}, which reduces to the convex hull of contact points when all contacts are coplanar and friction is sufficient.

In studies of this system, the same quantity $c + \dot{c}/\omega$ appeared under different names and motivations. In biomechanics, \citet{hof2005condition} identified it as the \emph{extrapolated center of mass}, a predictor of balance in human walking. In humanoid robotics, \citet{pratt2006capture} framed it as the \emph{capture point}, the location on the ground where the robot should step in order to come to a stop, while \citet{sugihara2009regulator} identified it when maximizing the basin of attraction of linear feedback controllers. \citet{takenaka2009real} introduced the eigendecomposition into convergent and divergent components of motion~(DCM), which \citet{englsberger2015dcm} later extended to three dimensions, and around which a line of walking controllers has since been developed~\citep{caron2019stair, mesesan2019dynamic, scianca2020mpc, dallard2025robust}.

Let us write Equation~\eqref{eq:lip} as a first-order linear system. For the sake of simplicity, we consider a single coordinate $c \in \mathbb{R}$ (the two axes are decoupled) and its corresponding state $x = \begin{bmatrix} c & \dot{c} \end{bmatrix}^\top$:
\begin{equation}
  \frac{\mathrm{d}}{\mathrm{d}t}
  \begin{bmatrix} c \\ \dot{c} \end{bmatrix}
  =
  \underbrace{\begin{bmatrix} 0 & 1 \\ \omega^2 & 0 \end{bmatrix}}_{A}
  \begin{bmatrix} c \\ \dot{c} \end{bmatrix}
  +
    \underbrace{\begin{bmatrix} 0 \\ -\omega^2 \end{bmatrix}}_B z .
  \label{eq:lip-first-order}
\end{equation}
The state matrix $A$ has eigenvalues $\pm\omega$. Its left eigenvectors $w_\xi = \begin{bmatrix} 1 & 1/\omega \end{bmatrix}$ and $w_\zeta = \begin{bmatrix} 1 & -1/\omega \end{bmatrix}$ define two coordinates $\xi := w_\xi \cdot x = c + \frac{\dot{c}}{\omega}$ and $\zeta := w_\zeta \cdot x = c - \frac{\dot{c}}{\omega}$, respectively called the \emph{divergent} and \emph{convergent} components of motion~\citep{takenaka2009real}. A direct calculation gives their decoupled dynamics as:
\begin{equation}
  \dot{\xi} = \omega (\xi - z),
  \qquad
  \dot{\zeta} = -\omega (\zeta - z) .
  \label{eq:dcm-ccm-dyn}
\end{equation}
Regulating the convergent component $\zeta$ is not necessary: its eigenvalue $-\omega$ is stable, therefore it tracks the input $z$ on its own. The DCM $\xi$, on the other hand, represents a quantity that \emph{must} be controlled, as it otherwise diverges $(|\xi| \to \infty)$ resulting in a robot fall.

\subsection{Koopman Eigenfunctions for Control Systems}

The Koopman operator framework opens the toolbox of linear systems theory to nonlinear systems~\citep{mauroy2020koopman}, and is increasingly applied in robot learning~\citep{shi2026koopman}. Introduced by \citet{koopman1931} for Hamiltonian systems, and extended to dissipative systems and data-driven analysis from the 2000s onward~\citep{mezic2021koopman}, it represents a nonlinear system $\dot{x} = f(x)$ by a \emph{linear} operator acting on observable functions $g : \mathcal{X} \to \mathbb{C}$ rather than on states $x \in \mathcal{X}$. In discrete time, where the flow map $F: x_k \mapsto x_{k+1}$ of $f$ describes the evolution of the system over a timestep $\delta t$, the Koopman operator $\mathcal{K}_F$ is defined by:
\begin{align}
    \mathcal{K}_F g = g \circ F, \ \textrm{ that is, } \ \forall k \in \mathbb{N},\ (\mathcal{K}_F g)(x_k) = g(x_{k+1}).
\end{align}
In continuous time, its infinitesimal generator $\mathcal{L}_f$, when it exists, is the Lie derivative along $f$: $(\mathcal{L}_f g)(x) = \nabla g(x) \cdot f(x)$, that is, the time derivative of $g$ along trajectories.

The price to pay for linearity is infinite dimension. Dynamic mode decomposition (DMD)~\citep{schmid2010dmd} and its extension to nonlinear observables (EDMD)~\citep{williams2015edmd} approximate $\mathcal{K}_F$ by least-squares regression on a finite dictionary of observables. Such dictionaries are rarely invariant under the dynamics, in which case the approximation exhibits spurious eigenfunctions that do not evolve as their eigenvalues predict~\citep[Section~1.4]{mauroy2020koopman}. Invariance holds, by contrast, for subspaces spanned by \emph{eigenfunctions} of the operator. An eigenfunction $\varphi$ with eigenvalue $\lambda \in \mathbb{C}$ satisfies:
\begin{equation}
    \mathcal{L}_f \varphi = \lambda \varphi,
    \ \textrm{ that is, } \
    \frac{\mathrm{d}}{\mathrm{d}t} \varphi(x(t)) = \lambda \varphi(x(t)).
  \label{eq:koopman-eig}
\end{equation}
Eigenfunctions are intrinsic coordinates in which the dynamics are, by construction, closed and linear.

The Koopman operator can be extended to handle control inputs in several ways: by lifting the sequence of future inputs into the state~\citep{korda2018mpc}, by indexing the operator with the current input~\citep[Section~1.5]{mauroy2020koopman}, or by identifying eigenfunctions of the \emph{autonomous} dynamics and modeling how the input acts on them~\citep{kaiser2021kronic, folkestad2020extended, korda2020optimal}. Let us consider the latter for a control-affine system $\dot{x} = f(x) + B(x) u$ such as a legged robot. By the chain rule, an eigenfunction $\varphi$ of the autonomous dynamics evolves as~\citep[Section~4.1]{kaiser2021kronic}:
\begin{equation}
    \frac{\mathrm{d}}{\mathrm{d}t} \varphi(x)
    = (\mathcal{L}_f \varphi)(x) + \nabla \varphi(x) \cdot B(x) u
    = \lambda \varphi(x) + \nabla \varphi(x) \cdot B(x) u .
  \label{eq:kronic}
\end{equation}
The linear part is inherited from the autonomous flow, while the control term $\nabla\varphi(x)\cdot B(x)$ is, in general, state-dependent, so that exact lifted models of controlled systems are bilinear rather than linear. It is state-independent when $\nabla \varphi \cdot B$ is constant, which will be the case in this work.

\subsection{The DCM as a Koopman Eigenfunction}

For a linear system $\dot{x} = Ax$, the Koopman eigenfunctions can be sought as linear functionals $\varphi(x) = w^{\top} x$ given by the left eigenvectors $w$ of $A$. Given the eigendecomposition~\eqref{eq:dcm-ccm-dyn} of the LIP, the convergent and divergent components of motion can be interpreted as the two linear Koopman eigenfunctions of the model: $\zeta$ for the stable eigenvalue $-\omega$, and $\xi$ for the unstable eigenvalue $+\omega$. The DCM dynamics~\eqref{eq:dcm-ccm-dyn} then corresponds to the eigenfunction-control equation~\eqref{eq:kronic}:
\begin{equation}
  \dot{\xi} = \underbrace{\omega}_{\lambda}\, \xi
  + \underbrace{(-\omega)}_{\nabla\xi \,\cdot\, B} z,
\end{equation}
with eigenvalue $\omega$, a linear observable $\varphi(x) = \xi(x) = w_\xi \cdot x$, and a state-independent control term. The standard DCM can thus be seen as a Koopman eigenfunction of the linear inverted pendulum model. But Koopman eigenfunctions can be identified for more general nonlinear systems, which brings us to the driving question of this study: can we use them to define DCMs applicable to locomotion control of nonlinear walking systems?

%===============================================================================

\section{Learning Divergent Components of Motion}
\label{sec:learning}

\subsection{Eigenfunction Search}

We now consider a legged robot balancing around an unstable equilibrium, modeled as a control-affine system $\dot{x} = f(x) + B(x) u$ whose dynamics $(f, B)$ are unknown. We represent its DCM by a function approximator $\xi_\theta$ with parameters $\theta$, and formulate the eigenfunction search as a gradient-based optimization over the pair $(\omega, \theta)$, where $\omega$ is the eigenvalue associated with $\xi_\theta$. We deliberately seek an unstable eigenvalue $\omega > 0$ that characterizes divergence away from the equilibrium. In this regard, our setting departs from prior Koopman-control studies such as~\citet{kaiser2021kronic} that targeted zero or low eigenvalues associated with conserved quantities and persistent dynamics.

Without a model, the state space on which to define $\xi_\theta$ is not known \emph{a priori}. We consider instead sequences $(y_k, u_k)_k$ of observations $y_k \in \mathcal{Y}$ and control inputs $u_k \in \mathcal{U} \subset \mathbb{R}^{n_u}$, from which we collect \emph{delay embeddings} with horizon $H$:
\begin{align}
    \mathcal{Z}_H & = \{ (y_{k-H}, u_{k-H}, \ldots, y_{k-1}, u_{k-1}, y_k) \} \subset (\mathcal{Y} \times \mathcal{U})^{H} \times \mathcal{Y}.
    \label{eq:delay-embedding}
\end{align}
Delay coordinates are a common substitute for the state in data-driven Koopman analysis~\citep{mezic2021koopman}. Observations $y_k$ may include joint angles, base orientation, joint torques or ankle force-torque measurements. We define the DCM as a map $\xi_\theta : \mathcal{Z}_H \to \mathbb{R}^{n_u}$ to the input space, so that it can be compared with $u$, and, inspired by the LIP DCM dynamics~\eqref{eq:dcm-ccm-dyn}, require that it satisfies:
\begin{equation}
  \dot{\xi}_\theta = \omega \, \xi_\theta - \omega\,u.
  \label{eq:dcm-ode}
\end{equation}
The first term makes $\xi_\theta$ an eigenfunction of the unforced dynamics with eigenvalue $\omega$. The second term constrains the control term $\nabla \xi \cdot B u$ of Equation~\eqref{eq:kronic} to be state-independent, and serves as our inductive bias. This choice is a first step rather than a requirement, and suffices for this study. We discuss more general input couplings in Section~\ref{sec:limitations}.

\subsection{DCM Dynamics Loss}

We assume a dataset $\mathcal{D}$ of measurement-action episodes $(y_k, u_k)_{k}$ collected on the robot under an existing controller. Integrating Equation~\eqref{eq:dcm-ode} under a zero-order hold of the input $u = u_k$ over a sampling period $\delta t = t_{k+1} - t_k$ yields the discrete-time DCM recursion:
\begin{equation}
  \xi_{k+1} = e^{\omega \delta t} \, \xi_k + \big(1 - e^{\omega \delta t}\big) \, u_k.
  \label{eq:dcm-discrete}
\end{equation}
This equation should hold over all steps $k$ of the dataset. Since our DCM observable $\xi_\theta$ is defined on delay embeddings $z_k \in \mathcal{Z}_H$, let us denote by $\mathcal{S}_{H+1}(\mathcal{D})$ the uniform sampling of measurement-action sequences $(y_{k-H}, u_{k-H}, \ldots, y_k, u_k, y_{k+1})$ spanning $H+1$ transitions within a single episode of $\mathcal{D}$. Each such sequence defines a pair $(z_k, z_{k+1})$ of consecutive embeddings, and we can enforce Equation~\eqref{eq:dcm-discrete} over $\mathcal{D}$ by minimizing its squared residual:
\begin{equation}
    \ell_{\mathrm{DCM}}^{(0)} = \mathbb{E}_{(z_k, z_{k+1}) \sim \mathcal{S}_{H+1}(\mathcal{D})} \left[ \left\| \xi_\theta(z_k) - e^{-\omega \delta t} \, \xi_\theta(z_{k+1}) - \left(1 - e^{-\omega \delta t}\right) \, u_k \right\|^2 \right],
  \label{eq:loss-dcm-td0}
\end{equation}
where $u_k$ is part of $z_{k+1}$. We refer to this loss as the one-step DCM loss.

\subsection{CoM Prediction Loss}

The DCM loss admits trivial minima at $\omega = 0$ with constant $\xi_\theta$, e.g., $(\omega, \xi_\theta) \equiv (0, 0)$, to which gradient descent may converge, even in the ideal LIP model. A general remedy to this issue is to learn a joint decoder that reconstructs states from encoded features, adding a corresponding reconstruction loss~\citep{korda2018mpc, yin2022embedding, bounou2024learning}. Rather than reconstructing a full state, which for a legged robot would typically involve generalized position and momentum coordinates, we exploit a second piece of LIP structure: the DCM anchors the (stable) CoM dynamics through:
\begin{equation}
  \dot{c} = \omega \, (\xi - c).
  \label{eq:com-ode}
\end{equation}
Under the zero-order hold $u = u_k$, we already know the analytical solution $\xi(t)$ of Equation~\eqref{eq:dcm-ode}, which Equation~\eqref{eq:dcm-discrete} gives at $t = t_k + \delta t$. Substituting it into Equation~\eqref{eq:com-ode} and integrating over $[t_k, t_{k+1}]$ yields an exact formula for the discrete-time CoM recursion:
\begin{equation}
  c_{k+1} = e^{-\omega \delta t} \, c_k + \big(1 - e^{-\omega \delta t}\big) \, \tfrac{\xi_k + \xi_{k+1}}{2}.
  \label{eq:com-discrete}
\end{equation}
We further assume that the measurements $y_k$, and by extension the embeddings $z_k$, allow estimating the CoM position $c_k = c(z_k)$ with respect to ground contact points. This allows us to add a one-step CoM prediction loss to our optimization problem:
\begin{equation}
    \ell_{\mathrm{CoM}}^{(0)} = \mathbb{E}_{(z_k, z_{k+1}) \sim \mathcal{S}_{H+1}(\mathcal{D})} \Big[ \big\| c_{k+1} - e^{-\omega \delta t} \, c_k - \big(1 - e^{-\omega \delta t}\big) \, \tfrac{\xi_\theta(z_k) + \xi_\theta(z_{k+1})}{2} \big\|^2 \Big].
  \label{eq:loss-com-td0}
\end{equation}
This loss penalizes trivial minima, as its residual collapses to $c_{k+1} - c_k$ irrespective of $\xi_\theta$ when $\omega \to 0$, leaving a non-vanishing floor $\mathbb{E}[\|c_{k+1} - c_k\|^2] > 0$. We can then combine it with the DCM loss as a weighted sum $\ell^{(0)}_\text{DCM} + w_\text{CoM} \, \ell^{(0)}_\text{CoM}$ with $w_\text{CoM} > 0$. We verify on the LIP model in Appendix~\ref{appendix:lip-eval} that increasing $w_\text{CoM}$ mitigates convergence to a trivial minimum.

\subsection{Multi-step Loss}

While the weighted one-step loss $\ell^{(0)}_\text{DCM} + w_\text{CoM} \, \ell^{(0)}_\text{CoM}$ is sufficient to recover the exact DCM of a LIP model, we found empirically that it tends to converge to overly low eigenvalues $\omega$ on real-robot data. To mitigate this effect, we evaluate the DCM observable against multi-step predictions. Defining $\gamma := e^{-\omega \delta t} \in (0, 1)$, we can rewrite the DCM recursion~\eqref{eq:dcm-discrete} as:
\begin{equation}
    \xi_k = (1 - \gamma) u_k + \gamma \, \xi_{k+1}.
  \label{eq:dcm-td-form}
\end{equation}
Coordinate by coordinate, this equation has the form of the Bellman equation of a Markov reward process, with value $\xi_k$, (scaled) reward $(1 - \gamma) u_k$, and discount factor $\gamma$. On a chunk of $m + 1$ consecutive embeddings $(z_k, \ldots, z_{k+m})$, iterating Equation~\eqref{eq:dcm-td-form} $n$ times yields the $n$-step DCM residual, that is, the difference between the value $\xi_\theta(z_k)$ and its $n$-step return:
\begin{equation}
    \delta^n_\text{DCM} := \xi_\theta(z_k) - \gamma^n \, \xi_\theta(z_{k+n}) - (1 - \gamma) \sum_{j=0}^{n-1} \gamma^j \, u_{k+j}.
    \label{eq:n-step-dcm-residual}
\end{equation}
Similarly, iterating the CoM recursion~\eqref{eq:com-discrete} forward yields the $n$-step CoM residual $\delta^n_\text{CoM} := c_{k+n} - \gamma^n \, c_k - \frac{1 - \gamma}{2} \sum_{j=0}^{n-1} \gamma^{n-1-j} \, ( \xi_\theta(z_{k+j}) + \xi_\theta(z_{k+j+1}) )$. We weight the squared $n$-step residuals with the geometric coefficients $(1 - \lambda) \lambda^{n-1}$ of the $\lambda$-return~\citep[Chapter~12]{sutton2018reinforcement}:
\begin{equation}
    \ell^{(\lambda)}_\text{CM} := \mathbb{E}_{(z_k, \ldots, z_{k+m}) \sim \mathcal{S}_{H+m}(\mathcal{D})} \! \left[ \sum_{n=1}^{m} (1 - \lambda) \, \lambda^{n-1} \, \left\|\delta^n_\text{M}\right\|^2 \right], \text{ where } \text{CM} \in \{\text{DCM}, \text{CoM}\},
  \label{eq:loss-lambda}
\end{equation}
and combine them into our final training objective $\ell^{(\lambda)} = \ell^{(\lambda)}_\text{DCM} + w_\text{CoM} \, \ell^{(\lambda)}_\text{CoM}$. Note that, while Equation~\eqref{eq:loss-lambda} weights $n$-step returns as in a (truncated) $\lambda$-return, our approach differs from TD($\lambda$) in two ways: the discount $\gamma(\omega)$ is a function of the learned eigenvalue $\omega$, and gradients also flow through the bootstrapped values $\xi_\theta(z_{k+n})$. When $\lambda = 0$, only the one-step residuals contribute, and we recover the one-step losses~\eqref{eq:loss-dcm-td0} and~\eqref{eq:loss-com-td0}.

%===============================================================================

\section{Application to Legged Balancing}
\label{sec:application}

\begin{figure}[th]
  \centering
  \begin{minipage}[c]{0.55\textwidth}
    \centering
    \small
    \begin{tabular}{l|l}
        \hline
        Delay-embedding horizon $H$ & $10$ samples \\
        Initial frequency $\omega_0$ & $5.0$~rad/s \\
        CoM-loss weight $w_\text{CoM}$ & $1.0$ \\
        Adam learning rate & $3 \times 10^{-4}$ \\
        Batch size $b$ & $1024$ chunks \\
        Validation fraction & $10\%$ \\
        Training duration & $100$ epochs \\
        Number of lambda steps $H_\lambda = 1/(1{-}\lambda)$ & $10$ \\
        Chunk length $m$ & $5 H_\lambda$ \\
        \hline
    \end{tabular}
  \end{minipage}\hfill
  \begin{minipage}[c]{0.40\textwidth}
    \centering
    \includegraphics[height=12em]{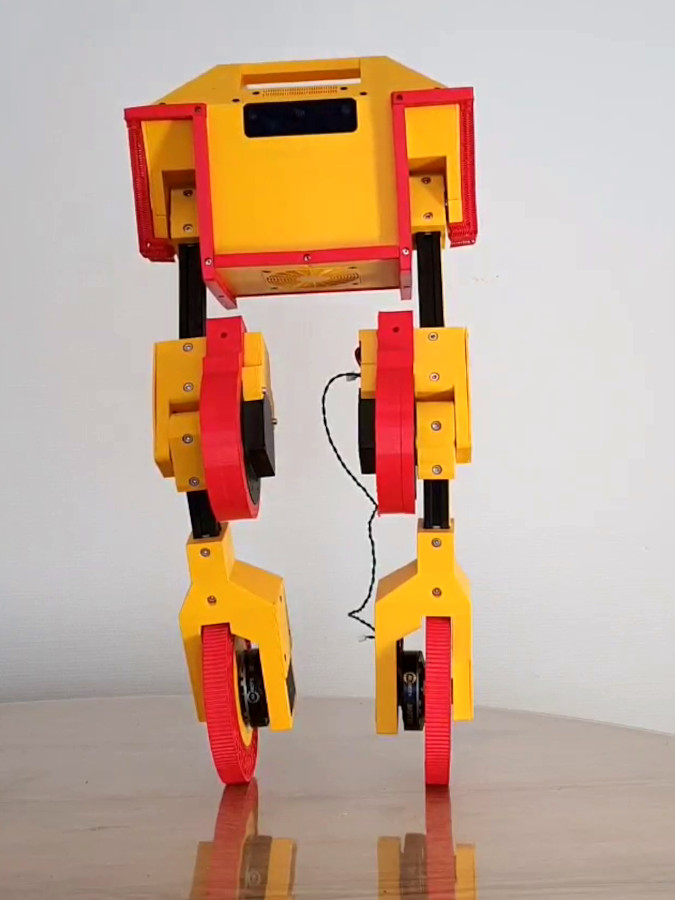}
  \end{minipage}
  \caption{Left: hyperparameter values used to train Koopman DCMs. Right: Cookie wheeled biped robot on which hardware experiments were carried out. Without abduction-adduction joints, the robot cannot move its CoM laterally without tilting its trunk, which breaks the assumptions of the linear inverted pendulum for which DCMs were originally derived.}
  \label{fig:cookie}
\end{figure}

We now instantiate the Koopman DCM learning pipeline on the lateral dynamics of the Cookie wheeled biped (Figure~\ref{fig:cookie}, right) from the Upkie open-hardware project~\citep{upkie}, while the sagittal dynamics are balanced by a separate model predictive controller~\citep{bambade2025proxqp}. The robot has three actuators per leg: two flexion-extension hip and knee joints, followed by a wheel in contact with the ground. Since the robot has no abduction-adduction joint, the lateral CoM position can only be controlled indirectly, through differential leg extension.

We selected the input as the lateral ZMP $u = z \in [-1, 1]$, normalized to $\pm 1$ at the wheel contacts, and collected closed-loop training data under a proportional DCM feedback law at the ZMP~\citep{pratt2006capture}:
\begin{equation}
    z = z_\text{ref} - \big(1 + k_p / \omega\big) \, \big( \xi_\text{ref} - \xi_\text{obs} \big).
    \label{eq:dcm-feedback}
\end{equation}
This commanded ZMP then drove a force-difference control loop~\citep{kajita2010biped} on leg extensions, which also takes as input the ZMP measured from joint torques by a quadratic-programming estimator~\citep{xinjilefu2014iros}. We used $k_p = 1$ rad/s in all experiments. This control law can use any pair $(\omega, \xi_{\text{obs}})$ of natural frequency and observed DCM: during data collection, we used the LIP frequency $\omega_{\mathrm{LIP}} = \sqrt{g/h}$ (recomputed at each control cycle from the instantaneous CoM height $h$) and the LIP DCM $\xi_{\mathrm{LIP}} = c + \dot{c} / \omega_{\mathrm{LIP}}$, whereas with a Koopman DCM we switched to the learned pair $(\hat{\omega}, \xi_{\theta})$.

\subsection{Data Collection}

The dataset $\mathcal{D}$ consists of one hour of closed-loop data collected on the real robot under the DCM feedback controller~\eqref{eq:dcm-feedback}, in $10$-second chunks tracking sinusoidal ZMP references $u^\text{ref}_t = o + A \sin(2\pi t / T)$, with parameters drawn uniformly from amplitudes $A \in \{0, 0.1, 0.2, 0.4, 0.8\}$, offsets $o \in \{0, \pm 0.2, \pm 0.5\}$, and periods $T \in \{0.3, 0.7, 1.0, 1.2, 1.4\}$~s. These distributions include stationary ($20\%$ of the data) and infeasible ($8\%$) references. A new episode starts whenever the robot falls or the user triggers a manual reset.

The CoM prediction loss~\eqref{eq:loss-com-td0} requires an estimate of the CoM $c_k$ at every timestep. At every control cycle, we estimated the floating-base configuration by differential inverse kinematics from joint encoders, IMU base orientation, and rolling-contact tasks at the two wheels. The CoM position ${}^{A}c_k \in \mathbb{R}^3$ follows from floating-base and joint-angle coordinates by forward kinematics. We express it in a local frame $A$ centered at the midpoint between the two wheel contacts and aligned with the robot's heading, so that its coordinates are invariant to the robot's pose in the inertial frame. The measurement vector is finally $y_k = (q_k, {}^{A}c_k, c_k) \in \mathbb{R}^{10}$, where $q_k \in \mathbb{R}^6$ is the vector of actuated joint angles and $c_k \in \mathbb{R}$ is the lateral CoM coordinate entering Equation~\eqref{eq:loss-com-td0}, normalized like the ZMP coordinate so that $c_k = +1$ above the left wheel and $c_k = -1$ above the right one. With $H = 10$, $n_y = 10$ and $n_u = 1$, the delay embedding $z_k$ has dimension $n_z = (n_y + n_u) \, H + n_y = 120$.

\subsection{Training}
\label{sec:training}

We represent the DCM observable $\xi_\theta$ by a two-hidden-layer multilayer perceptron of width $128$ with $\tanh$ activations, which maps the delay embedding $z_k$ to a scalar lateral coordinate ($n_u = 1$). The eigenvalue is parametrized by its logarithm $\log \omega \in \mathbb{R}$ to enforce $\omega > 0$. We train $\xi_\theta$ jointly with $\log\omega$ on $N = 330\,000$ transitions, that is, $55$~minutes of real-robot data at $100$~Hz ($\delta t = 10$~ms), holding out a random $10\%$ of the chunks for validation. At each iteration, we sample from the training chunks a mini-batch of $b$ chunks of $m + 1$ consecutive embeddings, each within a single episode, build the multi-step residuals $\delta^n_\text{DCM}$ and $\delta^n_\text{CoM}$ of Equation~\eqref{eq:loss-lambda}, and update $(\theta, \log\omega)$ by one step of the Adam optimizer on the combined loss $\ell^{(\lambda)} = \ell^{(\lambda)}_\text{DCM} + w_\text{CoM}\, \ell^{(\lambda)}_\text{CoM}$. We set the chunk length to $m = 5 H_\lambda$ so that the truncation tail $\lambda^m$ carries less than $1\%$ of the geometric weight.

Figure~\ref{fig:cookie} (left) summarizes the hyperparameter values used in this study. Whereas one-step losses suffice to recover $\omega$ on the LIP model (Appendix~\ref{appendix:lip-eval}), on real-robot data the one-step loss failed to identify a non-trivial eigenvalue: training collapsed to $\hat{\omega} \approx 0.05$~rad/s from every initialization we tested, despite the CoM loss. With $\lambda > 0$, the multi-step loss avoids this collapse and converges to an eigenvalue that is reproducible across random seeds. This eigenvalue depends on the number of lambda steps, from $3.30$ to $4.20$~rad/s for $H_\lambda \in \{5, 10, 20\}$, while the tracking performances of the resulting controllers lie within measurement dispersion of each other (Appendix~\ref{appendix:lambda-sweep}). We use the middle value $H_\lambda = 10$ ($\lambda = 0.9$) throughout this study, with a delay-embedding horizon $H = 10$.

\subsection{Learned-LIP Ablation}
\label{sec:ablation}

Our pipeline learns two quantities at once: the eigenfunction $\xi_\theta$ and its eigenvalue $\hat{\omega}$. To test whether the learned eigenfunction brings anything beyond the frequency, we consider a \emph{learned-LIP} ablation that removes it and keeps the analytical LIP observable, which amounts to fitting a LIP model to the collected data. We denote by $\tilde{\omega}$ the resulting frequency, to distinguish it from the Koopman eigenvalue $\hat{\omega}$, and define the learned-LIP DCM as:
\begin{equation}
  \xi_{\tilde{\omega}}(z_k) \: = \: c_k + \frac{\dot{c}_k}{\tilde{\omega}},
  \qquad
  \dot{c}_k \: := \: \frac{c_k - c_{k-1}}{\delta t},
  \label{eq:learned-lip}
\end{equation}
where $\dot{c}$, which is not directly measured, is obtained by backward finite difference from $c_{k-1}$ and $c_k$, both available in $z_k$. We fit $\tilde{\omega}$ by minimizing the same objective $\ell^{(\lambda)}$ over the same training split, with the same hyperparameters (Figure~\ref{fig:cookie}, left), $\tilde{\omega}$ being the only trained parameter.

\subsection{Results}
\label{sec:results}

Training over the collected dataset converged to an eigenvalue $\hat{\omega} = 3.85$~rad/s for the Koopman DCM, and to a natural frequency $\tilde{\omega} = 6.29$~rad/s for the learned-LIP ablation, with standard deviations of $0.014$ and $0.02$~rad/s respectively across five random seeds. The two data-driven values fall on either side of the kinematic $\omega_{\mathrm{LIP}} \approx 5.0$~rad/s: learning the eigenfunction jointly with its eigenvalue pulls the latter below $\omega_{\mathrm{LIP}}$, whereas fitting the frequency alone pushes it above.

Walking controllers track a reference trajectory, the walking pattern, which prescribes the CoM or its derivatives, such as CoM-ZMP~\citep{kajita2003preview} or CoM-velocity tracking~\citep{hwangbo2019learning}. To compare DCM estimators, a CoM tracking error would include variations of the convergent component, while a DCM tracking error would be expressed in coordinates that differ from one candidate DCM to the next. We therefore compare controllers by the root mean square error (RMSE) $\sqrt{\mathbb{E}_t [ (u_t - u^\text{ref}_t)^2 ]}$ of their tracking of a ZMP reference $u^\text{ref}_t$, a quantity with the same physical meaning for all candidates.

In this evaluation, we close the feedback law~\eqref{eq:dcm-feedback} with each candidate DCM and track sinusoidal references $u^\text{ref}_t = A \sin(2\pi t / T)$ at zero offset. The three candidate DCMs are:
\begin{itemize}
    \item The \emph{kinematic} baseline $\xi_{\mathrm{LIP}} = c + \dot{c} / \omega_{\mathrm{LIP}}$ with $\omega_{\mathrm{LIP}} = \sqrt{g / h}$.
    \item The \emph{learned-LIP} ablation $\xi_{\tilde{\omega}}$ of Equation~\eqref{eq:learned-lip}, with the fitted frequency $\tilde{\omega}$.
    \item The \emph{Koopman DCM}, with the learned eigenvalue $\hat{\omega}$ and observable $\xi_\theta$.
\end{itemize}
Each trial lasts ten periods. We discard the first two as settling, and compute, over the remaining eight, the RMSE, tracking gain and phase lag of the commanded ZMP with respect to its reference.

\begin{table}[th]
  \centering
  \footnotesize
  \setlength{\tabcolsep}{2.5pt}
  \caption{Real-robot ZMP tracking at period $T = 1.2$~s for the kinematic (Kin.), learned-LIP (L-LIP) and Koopman DCMs: mean $\pm$ half-width of the $95\%$ confidence interval over $10$ trials per cell.}
  \label{tab:zmp-results}
    \begin{tabular}[t]{c|r@{\,$\pm$\,}lr@{\,$\pm$\,}lr@{\,$\pm$\,}l}
      \multicolumn{7}{c}{RMSE} \\
      \hline
      $A$ & \multicolumn{2}{c}{Kin.} & \multicolumn{2}{c}{L-LIP} & \multicolumn{2}{c}{Koopman} \\
      \hline
      $0.1$ & $0.12$ & $0.01$ & $0.13$ & $0.02$ & $0.12$ & $0.02$ \\
      $0.2$ & $0.16$ & $0.02$ & $0.15$ & $0.03$ & $\mathbf{0.09}$ & $0.02$ \\
      $0.3$ & $0.16$ & $0.03$ & $0.15$ & $0.04$ & $\mathbf{0.11}$ & $0.01$ \\
      $0.4$ & $0.21$ & $0.02$ & $0.19$ & $0.03$ & $\mathbf{0.17}$ & $0.02$ \\
      $0.5$ & $0.27$ & $0.01$ & $0.28$ & $0.07$ & $\mathbf{0.23}$ & $0.05$ \\
      $0.6$ & $0.34$ & $0.05$ & $0.29$ & $0.04$ & $\mathbf{0.27}$ & $0.03$ \\
      \hline
    \end{tabular}\hspace{0.8em}%
    \begin{tabular}[t]{r@{\,$\pm$\,}lr@{\,$\pm$\,}lr@{\,$\pm$\,}l}
      \multicolumn{6}{c}{Gain (\%)} \\
      \hline
      \multicolumn{2}{c}{Kin.} & \multicolumn{2}{c}{L-LIP} & \multicolumn{2}{c}{Koopman} \\
      \hline
      $39$ & $5$ & $42$ & $5$ & $\mathbf{82}$ & $4$ \\
      $54$ & $8$ & $54$ & $7$ & $\mathbf{86}$ & $4$ \\
      $66$ & $8$ & $66$ & $9$ & $\mathbf{88}$ & $3$ \\
      $60$ & $6$ & $60$ & $7$ & $\mathbf{76}$ & $4$ \\
      $60$ & $6$ & $58$ & $6$ & $\mathbf{76}$ & $8$ \\
      $54$ & $7$ & $61$ & $8$ & $\mathbf{81}$ & $8$ \\
      \hline
    \end{tabular}\hspace{0.8em}%
    \begin{tabular}[t]{r@{\,$\pm$\,}lr@{\,$\pm$\,}lr@{\,$\pm$\,}l}
      \multicolumn{6}{c}{Phase lag (deg)} \\
      \hline
      \multicolumn{2}{c}{Kin.} & \multicolumn{2}{c}{L-LIP} & \multicolumn{2}{c}{Koopman} \\
      \hline
      $\mathbf{3}$ & $10$ & $6$ & $7$ & $21$ & $7$ \\
      $8$ & $5$ & $\mathbf{3}$ & $7$ & $8$ & $14$ \\
      $7$ & $6$ & $\mathbf{3}$ & $8$ & $23$ & $3$ \\
      $19$ & $3$ & $\mathbf{12}$ & $8$ & $27$ & $3$ \\
      $15$ & $7$ & $\mathbf{4}$ & $13$ & $14$ & $9$ \\
      $8$ & $8$ & $\mathbf{4}$ & $7$ & $9$ & $5$ \\
      \hline
    \end{tabular}
\end{table}

Table~\ref{tab:zmp-results} reports the three metrics for amplitudes $A \in \{0.1, \ldots, 0.6\}$ at $T = 1.2$~s. The Koopman DCM achieves a higher tracking gain than both baselines at all amplitudes with non-overlapping confidence intervals. This larger gain can also be observed in ZMP tracking plots, as depicted in Figure~\ref{fig:zmp-results} and in the supplementary video (linked from the source code repository). It comes however with a larger phase lag. Meanwhile, the learned-LIP baseline has a lower mean phase lag than the others, although with overlapping confidence intervals. These results suggest in particular that Koopman DCMs learn different features than fitting a LIP model.

\begin{figure}[h]
  \centering
  \includegraphics[width=\columnwidth]{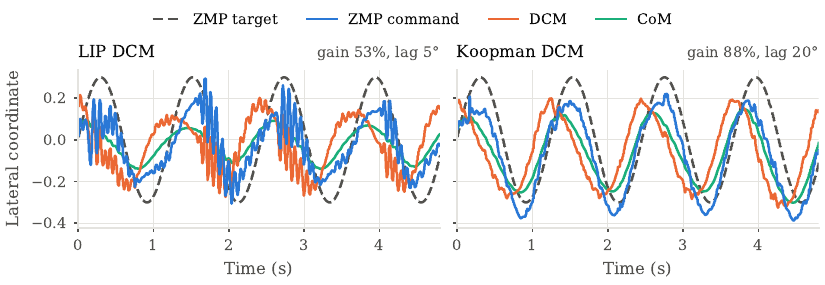}
    \caption{Real-robot ZMP tracking at amplitude $A = 0.3$ and period $T = 1.2$~s, from the recording of the supplementary video, under the LIP DCM (left) and the Koopman DCM (right). Signals are lateral coordinates normalized to $\pm 1$ at the wheel contacts, and each panel shows the DCM of its own controller. The commanded ZMP under the Koopman DCM follows the amplitude of the reference more closely (gain $88\%$ against $53\%$), with a larger phase lag ($20^\circ$ against $5^\circ$).}
  \label{fig:zmp-results}
\end{figure}

Evaluation references belong to the same family of sinusoids as the training data, with $T = 1.2$~s part of the training grid. We therefore also evaluated three additional periods $T \in \{0.5, 1.8, 2.0\}$~s at $A = 0.3$ with three trials each. We observed similar gains for the Koopman DCM, which remained higher than that of the kinematic baseline at the two longer periods ($86\%$ against $46\%$ at $T = 2.0$~s, $78\%$ against $52\%$ at $T = 1.8$~s) and closer to $100\%$ at $T = 0.5$~s ($103\%$ against $127\%$). Its RMSE was lower than that of the baseline at $T = 0.5$~s ($0.07$ against $0.14$) and comparable at the two longer periods ($0.19$ against $0.18$ at $T = 1.8$~s, $0.22$ against $0.19$ at $T = 2.0$~s).

\subsection{Learned DCM Analysis}
\label{sec:observable-analysis}

Since training and evaluation references come from the same family, the learned DCM could in principle infer the phase of the reference from its own action history rather than from the state of the robot. To check this, we replace parts of its input $z_k$ by values recorded at unrelated times (thus not useful for control but coming from the same distribution), and measure the resulting change in output $\mathrm{RMS}(\Delta \xi_\theta)$, in units of the standard deviation $\sigma(\xi_\theta)$ of the output. On the out-of-distribution data above, held out in full from training, scrambling past actions yields $\mathrm{RMS}(\Delta \xi_\theta) = 0.31\,\sigma(\xi_\theta)$, whereas scrambling measurements yields $1.51\,\sigma(\xi_\theta)$ ($0.32\,\sigma(\xi_\theta)$ and $1.47\,\sigma(\xi_\theta)$ on the validation split). The learned DCM thus relies mostly on measurements rather than on its own action history. Following a similar approach, we report its sensitivity to state-estimation noise in Appendix~\ref{appendix:noise}.

%===============================================================================

\section{Model Predictive Control over Learned DCMs}
\label{sec:mpc}

Viability is a property of states that is well-suited to the locomotion task. A state is viable as long as there exists a future evolution of the system, starting from it, that avoids fallen states. In the case of the linear inverted pendulum, \citet{sugihara2009regulator} identified that the input inequality constraint $C z \leq d$ translates directly to a viability constraint when applied to the linear DCM: the robot can regulate its future center-of-mass positions if and only if $C \xi \leq d$. Model predictive control (MPC) provides a way to take such constraints into account~\citep{wieber2008viability}.

To illustrate this, we formulate the lateral walking task as a constrained optimal control problem over a finite horizon of $N$ steps. The state is the (scalar) DCM $\xi_k$, the input is the (scalar) ZMP $u_k$, and the dynamics are the discrete-time DCM recursion of~\eqref{eq:dcm-discrete},
\begin{equation}
    \xi_{k+1} = e^{\omega \delta t} \, \xi_k + (1 - e^{\omega \delta t}) \, u_k,
    \label{eq:mpc-dynamics}
\end{equation}
where $\delta t$ is the control timestep and $\omega$ the eigenvalue of the unforced dynamics. Given a reference $u^\text{ref}_t$, as well as the DCM reference $\xi^\text{ref}_t$ obtained as the bounded solution of~\eqref{eq:dcm-ode} for this input, the cost is quadratic, with stage tracking on both $\xi$ and $u$ plus a terminal state cost:
\begin{equation}
    J = \sum_{k=0}^{N-1} \big[ Q \, (\xi_k - \xi^\text{ref}_k)^2 + R \, (u_k - u^\text{ref}_k)^2 \big] + Q_\text{T} \, (\xi_N - \xi^\text{ref}_N)^2 .
    \label{eq:mpc-cost}
\end{equation}
The horizon $N$ and weights $Q, R, Q_\text{T}$ are design parameters. Viability enters through hard inequality constraints on the DCM and control input:
\begin{align}
    |\xi_k| & \leq \xi_\text{max} \quad (k = 1, \ldots, N), \\
    |u_k| & \leq u_\text{max} \quad (k = 0, \ldots, N - 1) .
    \label{eq:mpc-constraints}
\end{align}
The input box constraint $|u_k| \leq u_\text{max} = 1$ is the support polygon ($C = [1, -1]^\top$ and $d = [1, 1]^\top$): the lateral ZMP must lie between the two contact boundaries, at $\pm 1$ in normalized units. The viability condition $|\xi_k| \leq \xi_\text{max}$ resulting from the input constraint is $\xi_\text{max}=1$, which we tighten to $\xi_\text{max} = 0.85$ to leave a safety margin. Together, \eqref{eq:mpc-dynamics}, \eqref{eq:mpc-cost}, and \eqref{eq:mpc-constraints} form a strictly convex quadratic program in the decision variables $(\xi_1, \ldots, \xi_N, u_0, \ldots, u_{N-1})$ given the initial state $\xi_0$.

We evaluated the MPC in lateral-balancing simulations in PyBullet~\citep{pybullet} on two robots: the Unitree G1 humanoid and the Cookie wheeled biped. The control pipeline reproduces the architecture of Section~\ref{sec:application}, with the DCM controller set to proportional feedback, the constrained MPC, or the same MPC without the viability constraint $|\xi_k| \leq \xi_\text{max}$. On each robot, all three controllers used the same DCM observable $(\hat{\omega}, \xi_\theta)$, trained from simulated PyBullet trajectories. Training resulted in $\hat{\omega} = 1.91$~rad/s for G1 and $\hat{\omega} = 4.73$~rad/s for Cookie. The MPC parameters were $N = 50$, $Q = R = Q_\text{T} = 1$. The resulting quadratic program was solved at each control cycle with ProxQP~\citep{bambade2025proxqp}.

\begin{table}[h]
  \centering
  \small
  \caption{Lateral-balancing evaluation in simulation: fall rate and ZMP tracking RMSE (mean $\pm$ half-width of the $95\%$ confidence interval), over $90$ trials per controller.}
  \label{tab:mpc-results}
  \begin{tabular}{l|cc}
  \multicolumn{3}{c}{G1 humanoid} \\
  \hline
  Controller & Fall rate & RMSE \\
  \hline
  Proportional DCM & $39\%$ ($35/90$) & $0.78 \pm 0.06$ \\
  MPC w/o viability & $50\%$ ($45/90$) & $0.72 \pm 0.03$ \\
  MPC & $\mathbf{0\%}$ ($0/90$) & $\mathbf{0.61} \pm 0.02$ \\
  \hline
  \end{tabular}\hspace{1em}%
  \begin{tabular}{cc}
  \multicolumn{2}{c}{Cookie wheeled biped} \\
  \hline
  Fall rate & RMSE \\
  \hline
  $28\%$ ($25/90$) & $0.15 \pm 0.02$ \\
  $33\%$ ($30/90$) & $0.14 \pm 0.02$ \\
  $\mathbf{17\%}$ ($15/90$) & $\mathbf{0.14} \pm 0.01$ \\
  \hline
  \end{tabular}
\end{table}

Table~\ref{tab:mpc-results} shows the results from sweeping sinusoidal ZMP references $u^\text{ref}_t = o + A \sin(2\pi t / T)$ over offsets $o \in \{0, 0.5\}$ and, on G1 and Cookie respectively, amplitudes $A \in \{0.3, 0.6, 0.9\}$ and $\{0.2, 0.4, 0.6\}$ and periods $T \in \{1.3, 1.6, 1.9\}$~s and $\{1.0, 1.2, 1.4\}$~s, repeated for five seeds ($90$ trials per controller). Falls were detected based on pelvis height on G1 and IMU tilt on Cookie. The larger amplitudes and the non-zero offset $o = 0.5$ push the reference DCM toward the edge of the support polygon, stressing the controllers' viability margin. On G1, the robot fell in $35/90$ trials under proportional DCM feedback and in $45/90$ under the MPC without viability constraint, whereas it did not fall under the MPC with active viability constraint. On Cookie, viability roughly halved MPC falls ($15/90$ against $30/90$) without eliminating them entirely.

%===============================================================================

\section{Limitations}
\label{sec:limitations}

The most consequential choice in this study was to fix the input coupling from the start: in our DCM dynamics~\eqref{eq:dcm-ode}, the input enters linearly and at the same rate $\omega$ as the state. In the general control-affine setting~\eqref{eq:kronic} the coupling can be any $\nabla\xi \cdot B(x)$, and exact lifted models are then bilinear rather than linear~\citep[Section~3.1]{strasser2026overview}. More general DCM dynamics have been studied in reduced models, notably the bilinear form $\dot{\xi} = \omega \xi - \lambda z$ in the variable-height inverted pendulum, where the height stiffness $\lambda$ enters multiplicatively~\citep{caron2019vhip}. Learning their data-driven counterparts is left to future work.

Our hardware evaluation was also restricted to lateral balance on a wheeled biped in double support. Stepping, and more challenging locomotion tasks, remain to be evaluated. In general, there is no a priori reason for the unstable Koopman spectrum to contain a single eigenpair of interest: Koopman eigenvalues form an algebra, and selecting among unstable eigenpairs remains an open question. Robots with several balance modes, such as quadrupeds that can stand on two legs, may exhibit multiple large unstable eigenvalues, which would call for learning eigenfunctions jointly.

%===============================================================================

\section{Conclusion}
\label{sec:conclusion}

We presented an approach that treats divergent components of motion as data-driven Koopman eigenfunctions, learned from closed-loop measurement-action histories without committing to a reduced model. Our pipeline recovers the analytical DCM of a LIP model, and on a real wheeled biped, learns features that improve ZMP tracking compared to reduced-model DCMs. This improvement requires no sim-to-real transfer, as DCMs are trained directly on robot data. Learned DCMs also provide a state on which model predictive control can enforce viability constraints. Beyond model-based control, they could serve as physically grounded observations or safety constraints for learning-based locomotion policies.

%===============================================================================

\clearpage
\acknowledgments{We thank Nicolas Perrin-Gilbert, Jean Ponce and Olivier Sigaud for their insightful feedback on this work, and Alexandre Mauroy and Igor Mezi\'c for pointing us to relevant work on Koopman operator theory. We also thank Area Chair V2Z4 and Reviewers Qgs9, Zch9 and rG4m for their constructive questions and comments on an earlier version of this manuscript.}

%===============================================================================

% no \bibliographystyle is required, since the corl style is automatically used
\bibliography{refs}

%===============================================================================

\appendix

\section{Evaluation on the Linear Inverted Pendulum Model}
\label{appendix:lip-eval}

We evaluate our learning pipeline on a linear inverted pendulum (LIP) model, an ideal setting with fewer confounding factors than real-robot data. Since the ground-truth eigenpair is known, this setting validates the training pipeline itself, while the evidence for learned DCMs on hardware is their closed-loop tracking performance (Section~\ref{sec:results}).

We select a ground-truth natural frequency $\omega^\star = \sqrt{g/h} = 5$ rad/s. We set the observation vector to $y_k = (c_k, \dot c_k)$, with $c$ the position of the center of mass in the inertial frame. We then collected a dataset of $200\,000$ closed-loop steps tracking a sinusoidal ZMP reference with period $T \in \{0.3, 0.5, 0.8, 1.2, 2.0\}$~s, amplitude (normalized in $[-1, 1]$) $A \in \{0.1, 0.2, 0.3, 0.4, 0.5\}$ and stationary offset $o \in \{-0.2, -0.1, 0, 0.1, 0.2\}$.

\begin{figure}[th]
  \centering
  \includegraphics[width=0.7\linewidth]{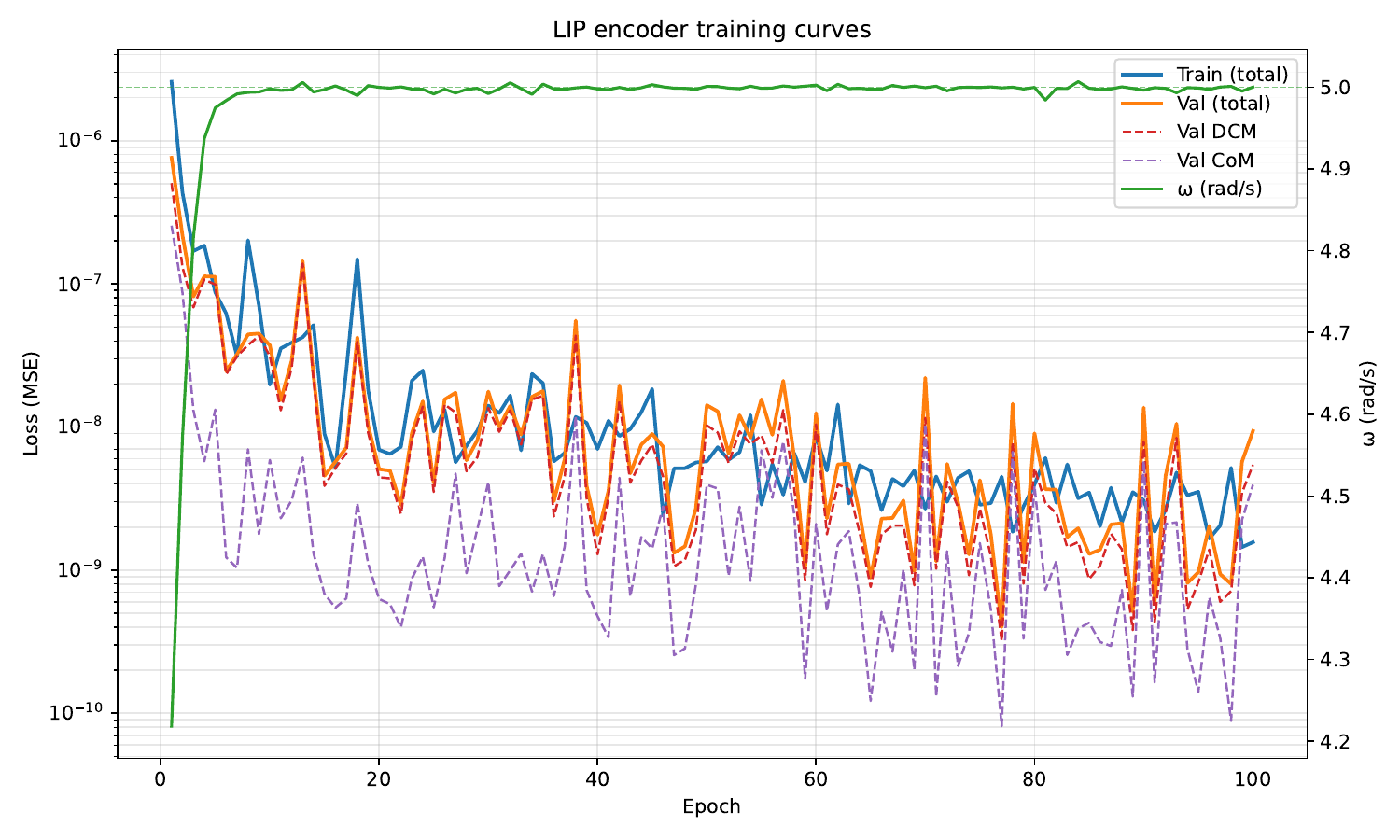}
  \caption{Training of a Koopman DCM estimator on a LIP model. Training and validation losses (solid), unweighted DCM and CoM sub-losses (dashed), and learned $\omega$ on the right axis (green).}
  \label{fig:lip-losses}
\end{figure}

These parameters are re-drawn every $2\,000$ steps uniformly at random from their corresponding sets. The DCM observable $\xi_\theta$ is a two-hidden-layer MLP of width $128$ with $\tanh$ activations, optimized jointly with $\log\omega$ by the Adam algorithm, with a learning rate of $10^{-3}$ and a batch size of $1024$, over $100$ epochs. The CoM-loss weight is set to $w_\text{CoM} = 1$ in this trial.

Figure~\ref{fig:lip-losses} shows a training run from $\omega_0 = 4.0$~rad/s, which converged to $\hat{\omega} = 4.9992$~rad/s (relative error $1.6 \!\times\! 10^{-4}$). The final validation residuals are $\ell^{(0)}_\text{DCM} = 5.4 \!\times\! 10^{-9}$ and $\ell^{(0)}_\text{CoM} = 3.9 \!\times\! 10^{-9}$, with both losses decreasing at similar rates over training. Training thus recovers the ground-truth eigenpair from this initialization; we examine below the role of the CoM residual in avoiding the trivial minimum $(\omega, \xi_\theta) \equiv (0,0)$ of the one-step DCM loss $\ell^{(0)}_\text{DCM}$.

\subsection{Weight of the CoM Prediction Loss}

We investigate the effect of the CoM prediction loss by sweeping the corresponding weight $w_\text{CoM} \in \{0.1, 1, 10, 100\}$, starting training from an initial guess $\omega_0 = 1.0$~rad/s close to the trivial minimum. We use three seeds for each value of the weight:

\begin{center}\small
\begin{tabular}{c|c|c}
\hline
$w_\text{CoM}$ & $\hat{\omega}$ (rad/s) & val.\ loss (MSE) \\
\hline
$0.1$  & $0.352 \pm 0.007$ & $7.35 \!\times\! 10^{-7}$ \\
$1$    & $4.975 \pm 0.003$ & $1.30 \!\times\! 10^{-8}$ \\
$10$   & $4.878 \pm 0.021$ & $7.46 \!\times\! 10^{-8}$ \\
$100$  & $3.800 \pm 0.051$ & $4.66 \!\times\! 10^{-6}$ \\
\hline
\end{tabular}
\end{center}

\begin{figure}[th]
  \centering
  \includegraphics[width=0.9\linewidth]{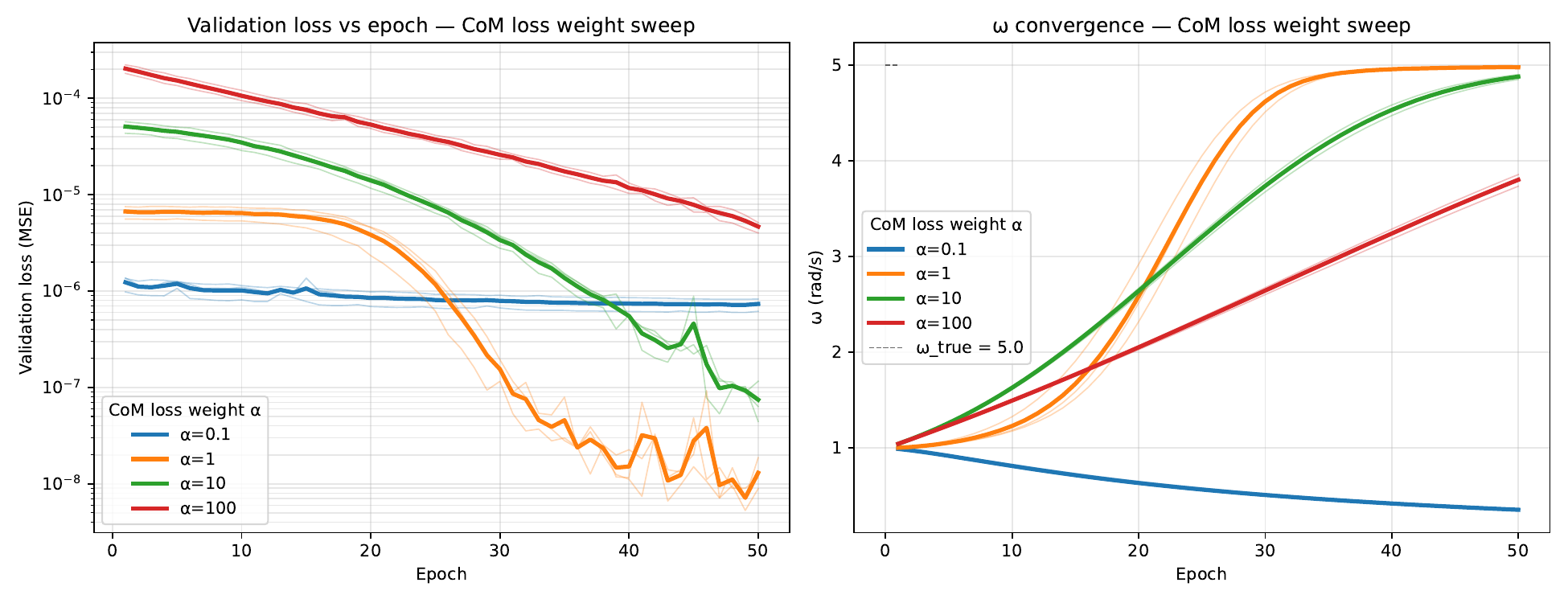}
  \caption{CoM-weight sweep at $\omega_0 = 1.0$~rad/s, three seeds each. Left: validation loss vs epoch; right: learned $\omega$ vs epoch. Only $w_\text{CoM} = 1$ recovers $\omega^\star$ from this initialization.}
  \label{fig:lip-com-sweep}
\end{figure}

Figure~\ref{fig:lip-com-sweep} shows three regimes. At $w_\text{CoM} = 0.1$, the DCM term dominates and $\hat{\omega}$ remains in the trivial $\omega \to 0$ basin. At $w_\text{CoM} = 1$, $\hat{\omega} = 4.975$~rad/s is within $0.5\%$ of $\omega^\star$, at the lowest validation loss. Finally, at $w_\text{CoM} \in \{10, 100\}$, the CoM term dominates and $\hat{\omega}$ approaches $\omega^\star$ more slowly, reaching $4.88$ and $3.80$~rad/s after $100$ epochs. The CoM prediction loss therefore regularizes training away from the trivial attractor, while too large a weight slows down convergence.

\subsection{Delay-embedding Horizon}

Table~\ref{tab:history-sweep} sweeps the delay-embedding horizon $H \in \{1, 3, 5, 10, 20\}$, with $10$ seeds per horizon, $50$ epochs over $50\,000$ closed-loop steps, and all other hyperparameters unchanged:

\begin{table}[ht]
  \centering\small
  \caption{Delay-embedding horizon $H$ on the LIP model: final $\hat{\omega}$ and validation loss (mean $\pm$ std over $10$ seeds).}
  \label{tab:history-sweep}
  \begin{tabular}{c|c|c}
  \hline
  $H$ & $\hat{\omega}$ (rad/s) & validation loss (MSE) \\
  \hline
  $1$  & $4.997 \pm 0.005$ & $2.37 \!\times\! 10^{-8}$ \\
  $3$  & $4.997 \pm 0.005$ & $6.02 \!\times\! 10^{-8}$ \\
  $5$  & $4.999 \pm 0.002$ & $1.90 \!\times\! 10^{-8}$ \\
  $10$ & $4.995 \pm 0.005$ & $2.79 \!\times\! 10^{-8}$ \\
  $20$ & $4.994 \pm 0.003$ & $4.12 \!\times\! 10^{-8}$ \\
  \hline
  \end{tabular}
\end{table}

% Table~\ref{tab:history-sweep} reports the aggregate statistics of eigenvalues and validation losses at the end of training
All five horizons recover $\hat{\omega}$ within $0.006$~rad/s of $\omega^\star = 5.0$~rad/s, with final validation losses of the same order of magnitude. On the LIP model, DCM estimation is therefore insensitive to $H$.
% \begin{figure}[ht]
%   \centering
%   \includegraphics[width=\linewidth]{figures/2026-06-05_112403_lip_eval_history_sweep_summary_cf340a3.pdf}
%   \caption{History-size sweep on the LIP, $10$ seeds per value. Left: final $\hat{\omega}$ vs $H$ (mean $\pm$ std; dashed line marks $\omega^\star = 5$~rad/s). Right: final validation loss vs $H$. The embedding horizon has no measurable effect on either quantity.}
%   \label{fig:lip-history}
% \end{figure}
This insensitivity is expected, as the observation vector $y_k = (c_k, \dot{c}_k)$ already contains both coordinates from which the eigenfunction $\xi = c + \dot{c}/\omega$ is built. It does not carry over to the real robot, whose measurement vector exposes no such pair of coordinates, and where we fix $H = 10$.

%===============================================================================

\section{Additional Real-robot Experiments}
\label{appendix:additional-experiments}

\subsection{Sensitivity to the Number of Lambda Steps}
\label{appendix:lambda-sweep}

The number of lambda steps $H_\lambda := 1 / (1 - \lambda)$ is the effective horizon of the geometric weights $(1 - \lambda) \lambda^{n-1}$ applied to the multi-step residuals, and sets the chunk length $m = 5 H_\lambda$. It is distinct from the delay-embedding horizon $H$, which counts the past measurement-action pairs in an embedding $z_k$, even though both equal $10$ in the configuration retained on Cookie. The DCM observables compared here all share $H = 10$ and differ only in $H_\lambda$.

We train DCM observables at $H_\lambda \in \{5, 10, 20\}$, all other hyperparameters being held fixed. Each value identifies a distinct natural frequency, reproducible across five random seeds:

\begin{center}\small
\begin{tabular}{c|c|c|c}
\hline
$H_\lambda$ & $\lambda$ & $m$ & $\hat{\omega}$ (rad/s) \\
\hline
$5$  & $0.8$  & $25$  & $3.299 \pm 0.015$ \\
$10$ & $0.9$  & $50$  & $3.855 \pm 0.014$ \\
$20$ & $0.95$ & $100$ & $4.196 \pm 0.006$ \\
\hline
\end{tabular}
\end{center}

The identified frequency increases monotonically with $H_\lambda$: weighting longer-range residuals more heavily drives the estimate further from the trivial minimum $(\omega, \xi_\theta) \equiv (0, 0)$. We close the loop with each of the three learned DCMs and measure ZMP tracking on a sinusoidal ZMP reference of amplitude $A$ and period $T$, reporting the root mean square error (RMSE) over the last two of at least ten periods:

\begin{table}[h]
  \centering
  \small
  \caption{ZMP tracking error on the real robot for three numbers of lambda steps $H_\lambda$, mean $\pm$ half-range across three sessions. All DCM observables use a delay-embedding horizon $H = 10$.}
  \label{tab:lambda-sweep}
  \begin{tabular}{c|cccc}
  \hline
  $A$ & kinematic & $H_\lambda = 5$ & $H_\lambda = 10$ & $H_\lambda = 20$ \\
  \hline
  $0.1$ & $0.11 \pm 0.06$ & $0.13 \pm 0.05$ & $0.08 \pm 0.05$ & $0.09 \pm 0.05$ \\
  $0.2$ & $0.12 \pm 0.01$ & $0.08 \pm 0.01$ & $0.05 \pm 0.01$ & $0.06 \pm 0.01$ \\
  $0.3$ & $0.20 \pm 0.06$ & $0.11 \pm 0.02$ & $0.11 \pm 0.02$ & $0.10 \pm 0.01$ \\
  \hline
  \end{tabular}
\end{table}

On three real-robot sessions at $T = 1.2$~s and $A \in \{0.1, 0.2, 0.3\}$, all three learned DCMs improve on the kinematic baseline. The lowest mean RMSE is attained at $H_\lambda = 10$ for $A \in \{0.1, 0.2\}$, and at $H_\lambda = 20$ for $A = 0.3$. At fixed amplitude $A = 0.3$ and periods $T \in \{0.8, 1.6\}$~s, $H_\lambda = 10$ attains the lowest RMSE at both periods ($0.05$ and $0.08$, against $0.14$ and $0.13$ for the kinematic baseline), while $H_\lambda = 20$ ($0.12$ and $0.14$) and $H_\lambda = 5$ ($0.07$ and $0.14$) both degrade at the longer period.

The differences between the three values are nonetheless within measurement dispersion: at every amplitude in Table~\ref{tab:lambda-sweep}, the three learned DCMs lie within each other's half-ranges. These data separate the learned DCMs from the kinematic baseline, but do not support a preference for any particular $H_\lambda$ on tracking grounds. We use the middle value $H_\lambda = 10$ throughout this study.

\subsection{Sensitivity to State-estimation Noise}
\label{appendix:noise}

We checked the sensitivity of the learned DCM of Section~\ref{sec:observable-analysis} to state-estimation noise. From data where the robot holds still, we estimate the standard deviation of high-frequency CoM residuals at about $0.5$~mm, sample noise from these segments, and add it to the delay embeddings of the validation split. The output then moves by $\mathrm{RMS}(\Delta \xi_\theta) = 0.09\,\sigma(\xi_\theta)$, which corresponds after feedback to an RMS change of $0.03$ in the commanded ZMP, against a standard deviation of $0.39$ for the commanded ZMP itself. We therefore expect state-estimation noise to have some effect on tracking performance, but to be less limiting than other factors (such as the bandwidth of the force-difference control loop).

\end{document}